\documentclass[letterpaper]{article} 
\usepackage{aaai23}  
\usepackage{times}  
\usepackage{helvet}  
\usepackage{courier}  
\usepackage[hyphens]{url}  
\usepackage{graphicx} 
\usepackage{natbib}  
\usepackage{caption} 
\usepackage{placeins}

\usepackage{algorithm}
\usepackage{algorithmic}

\usepackage{newfloat}
\usepackage{listings}
\DeclareCaptionStyle{ruled}{labelfont=normalfont,labelsep=colon,strut=off} 
\floatstyle{ruled}
\newfloat{listing}{tb}{lst}{}
\floatname{listing}{Listing}
\usepackage[utf8]{inputenc} 
\usepackage{url}            
\usepackage{booktabs}       
\usepackage{amsfonts}       
\usepackage{nicefrac}       
\usepackage{microtype}      
\usepackage{xcolor}         
\usepackage{mathtools}      
\usepackage{bbm}            
\usepackage{amsthm}         
\usepackage{amsmath}
\usepackage[symbol]{footmisc}

\usepackage{bm}
\DeclarePairedDelimiter{\abs}{|}{|}
\DeclarePairedDelimiter{\brac}{[}{]}
\DeclarePairedDelimiter{\braces}{\{}{\}}
\DeclarePairedDelimiter{\paren}{\lparen}{\rparen}

\DeclarePairedDelimiter\ang{<}{>}

\newcommand{\mc}[1]{\ensuremath \mathcal{#1}}

\newcommand{\AUC}[0]{\ensuremath \textup{AUC}}

\newcommand{\real}[0]{\mathbb{R}}
\newcommand{\extReal}[0]{\mathbb{R}}
\newcommand{\expt}[0]{\mathbb{E}}
\newcommand{\indc}[2]{\mathbb{I}_{#1}\paren*{#2}}
\newcommand{\dirac}[2]{\delta_{#1}\paren*{#2}}
\newcommand{\oddsRatio}[0]{\textup{OR}}

\newcommand{\xset}[0]{\mc{X}}
\newcommand{\yset}[0]{\mc{Y}}

\newcommand{\gset}[0]{\mc{G}}

\newcommand{\pset}[0]{\mc{P}}
\newcommand{\DSL}[0]{\mc{L}}
\newcommand{\DSU}[0]{\mc{U}}
\newcommand{\group}[1]{\ensuremath \bar{#1}}

\newcommand{\lbl}[1]{\tilde{#1}}
\newcommand{\lblhat}[1]{\check{#1}}
\newcommand{\ul}[1]{#1}

\newcommand{\Ind}[2]{\mathbb{I}_{#1}\paren*{#2}}
\newcommand{\Indtwo}[3]{\mathbb{I}_{#1}^{#2}\paren*{#3}}
\newcommand{\ccInv}[0]{CC-invariance}
\newcommand{\cpccInv}[0]{PCC-invariance}
\newcommand{\cl}[1]{\ensuremath {#1}'}
\newcommand{\andd}[0]{\ \&\ }

\newcommand{\seniorauthor}{\footnote{These authors should be considered as senior authors.}}

 \newtheorem{thm}{Theorem}

\title{Leveraging Structure for Improved Classification of Grouped Biased Data}
\author {
    Daniel Zeiberg\equalcontrib, 
    Shantanu Jain\equalcontrib\seniorauthor, 
    Predrag Radivojac\textsuperscript{\textdagger} 
}
\affiliations {
    Khoury College of Computer Sciences\\
    Northeastern University\\
    Boston, MA 02115, U.S.A.
}

\begin{document}

\maketitle

\begin{abstract}
We consider semi-supervised binary classification for applications in which data points are naturally grouped (e.g., survey responses grouped by state) and the labeled data is biased (e.g., survey respondents are not representative of the population). The groups overlap in the feature space and consequently the input-output patterns are related across the groups. 
To model the inherent structure in such data, we assume the partition-projected class-conditional invariance across groups, defined in terms of the group-agnostic feature space. We demonstrate that under this assumption, the group carries additional information about the class, over the group-agnostic features, with provably improved area under the ROC curve. Further assuming invariance of partition-projected class-conditional distributions across both labeled and unlabeled data, we derive a semi-supervised algorithm that explicitly leverages the structure to learn an optimal, group-aware, probability-calibrated classifier, despite the bias in the labeled data. Experiments on synthetic and real data demonstrate the efficacy of our algorithm over suitable baselines and ablative models, spanning standard supervised and semi-supervised learning approaches, with and without incorporating the group directly as a feature.
\end{abstract}

\section{Introduction}
Overcoming the problems of learning from biased data is among the most important challenges towards widespread adoption 
of data-driven technologies \cite{Schwartz2021}. Training machine learning models on biased data may lead to an unacceptable performance deterioration when deployed in the real world, and even more pernicious effects manifest as issues of fairness, when machine learning algorithms systemically lead to worse outcomes for a group of individuals \cite{Mehrabi2022}. However, despite the risks, it is often necessary to train models on biased data as it typically contains signal---in the context of classification, the labeled data may be biased, but it also contains class labels necessary for learning input-output patterns. Correcting for bias is particularly challenging since the mechanisms leading to it are often hidden in the complexities of the data generation process and are difficult to model or evaluate accurately \cite{storkey2009training}. For example, health care data may be biased due to issues of privacy or self-selection, and disease variant databases may be biased towards well-studied or easy-to-study genes and diseases \cite{Stoeger2018}.

A training dataset is biased if 
the data on which the model is applied cannot be interpreted to be drawn from the same probability distribution. Correcting for bias often relies on assuming some bias model that captures the relationship between biased and unbiased data distributions \cite{storkey2009training}, and then estimating bias correction parameters from the unbiased data \cite{heckman1979sample, cortes2008sample}. Thus, 
bias correction comes under the semi-supervised framework, where unlabeled data is considered to be unbiased and is used to correct the biases arising from training on labeled data.

Covariate shift and 
label shift are the two most well studied bias assumptions in 
classification \cite{storkey2009training}. For $x$ and $y$ representing input and 
output variables, covariate shift assumes that, though the distribution of inputs, $p(x)$, may be different in the labeled and unlabeled data, the distribution of the output at a given input, $p(y|x)$, stays constant. In theory, there is no need for bias correction here, as a nonparametric model trained to learn $p(y|x)$ is not affected by the bias \cite{Rojas1996}. Under label shift, instead of the invariance of $p(y|x)$, the invariance of class-conditionals $p(x|y)$ is assumed. Consequently, the difference in $p(y|x)$ from labeled to unlabeled data is attributed to the change in class priors $p(y)$. Here, the correction of bias relies on an elegant solution of using the model trained on labeled data to estimate the unbiased class priors from unlabeled data \cite{Vucetic2001, mlls}.

Though the bias under covariate and label shift can be effectively controlled, the assumptions are too restrictive for many real-world datasets where neither $p(y|x)$ nor $p(x|y)$ are invariant. 
We therefore introduce a more flexible bias assumption, partition-projected class-conditional invariance (\cpccInv), that allows both $p(y|x)$ and $p(x|y)$ to vary between labeled and unlabeled data. The assumption relies on the existence of a natural partitioning of the input feature space \cite{Chapelle2005}, where instead of assuming the invariance of the standard class-conditional distributions, we assume invariance of class-conditional distributions restricted to clusters of the data. 

In addition to bias, real-world data often has inherent structure, which may be leveraged for improved learning. Such structure might come from existing features, domain knowledge or additional metadata. For example, census data from different states might be closely related in that input-output patterns learned on Massachusetts can be useful to make predictions on California. General purpose machine learning methods capture such relationships to an extent; however, when such relationships are explicitly modeled by a learner, significant performance gains may be achieved. Furthermore, such structure, when exploited, may also counter the presence of bias. For example, if the labeled data has an underrepresentation of low income households from California, the patterns learned on the low income households from Massachusetts can be exploited.

Here, we consider structured data in domains where objects appear in naturally occurring groups; e.g., individuals grouped by state and variants grouped by gene. Similar to the bias model, our structure assumption is also expressed as PCC-invariance 
across the groups. The flexibility of this approach allows each group to have a different class-conditional distribution. We show that, under this assumption, a group-aware classifier that incorporates the group information, along with other features, has provably better performance than a group-agnostic classifier. However, the straightforward approach to incorporate the group information as a one-hot encoding is often sub-optimal in practice. Our proposed approach that exploits \cpccInv\ across the groups and between labeled and unlabeled data performs better on synthetic and real datasets as compared to the baseline group-aware and group-agnostic classifiers as well as other ablative models.

\section{Problem Formulation}
\label{sec:probform}
In the context of binary classification, let each object in the population of interest have a representation in the feature space $\xset$ and a class label in $\yset = \braces*{0,1}$. Additionally, let each object in the population belong to a distinct group, identified by a group index in $\gset =\{1,2, \ldots, G\}$. The objects from different groups may overlap in the group-agnostic feature space $\xset$; i.e., their $\xset$ representations can be arbitrarily close and even coincide. Finally, let $x \in \xset, y\in \yset$ and $g \in \gset$ be the random variables giving the group-agnostic representation, class label and group of an object in the population and $p(x,g,y)$ be the joint distribution of the variables over the population. 

\textbf{CC-Invariance}: Under the assumption of class-conditional invariance across groups (\ccInv), also referred to as label shift in the domain adaptation literature \cite{labelShiftUnified}, the class-conditional distributions of $x$ given $y$ are independent of $g$; i.e.,
$$p(x|y,g) = p(x|y).$$ In other words, the distribution of the input representations coming from the positives ($y=1$) or the negatives ($y=0$) is the same in all groups. However, the groups may differ in the proportion of positives and negatives, so it is possible that $p(y) \neq p(y|g=i) \neq p(y|g=j)$ for some $i\neq j$. Intuitively, the group index does not contain any information about the positive ($p(x|y=1)$) and negative ($p(x|y=0)$) class-conditional distributions, but it still contains information about the class label, as the proportions of positives and negatives differ across the groups. Consequently, in theory, a classifier trained to learn the group-aware posterior, $p(y=1|x,g)$, would be better at predicting the class label.

\textbf{PCC-Invariance}: 
The \ccInv\ assumption might be inadequate for datasets with more complex structure, where the class-conditional distributions differ across groups; i.e., $p(x|y,g=i) \neq p(x|y, g=j)$, for $i \neq j$. If the distributions change arbitrarily across groups, there is no special structure present in the data to be exploited by a learning algorithm. However, many datasets have a clustering structure that can be of use \cite{Chapelle2005}. We therefore assume the partition-projected class-conditional invariance across groups (\cpccInv), defined in terms of a partition of $\xset$. 

Let $\braces*{\mc{X}_k}_{k=1}^{K}$ be a family of subsets of $\xset$ that partitions it. We refer to the subsets as clusters. Let $\pi:\xset \rightarrow \pset = \braces*{1, 2, \ldots, K}$ be a function that maps each $x \in \xset$ to the index of the cluster it belongs to; i.e., for $x \in \xset_k$, $\pi(x) = k$.  Formally, \cpccInv\ is defined by the condition
\begin{align*}
p(x|y,g,\pi(x)) &= p(x|y,\pi(x)),  
\end{align*}
where $p(x|y,\pi(x))$ is the class-conditional distribution projected on the partition containing $x$.  Observe that the \ccInv\ is a special case of \cpccInv, when $K=1$. 

For an unambiguous exposition, we refer to the standard class-conditional, $p(x|y)$, simply as the class-conditional and the class-conditional for a group, $p(x|y,g)$,  as the group class-conditional. As a consequence of \cpccInv, each  group class-conditional can be expressed as a mixture with the partition-projected class-conditionals as components. The partition proportions in the positives or negatives of the group correspond to the mixing weights. Formally, using $\pi$ also as a random variable giving the cluster index, we have
\begin{align*}
p(x|y,g) &= \sum_{k=1}^K p(\pi = k|y,g) p(x|y,\pi=k)  \\
        &= p(\pi(x)|y,g) p(x|y,\pi(x)). 
\end{align*}
The last expression simplifies the group $g$ class-conditional at $x$ as a product of the class-conditional projected on the partition containing $x$ and the proportion of that partition among the positives or negatives of the group. Note that, though the same components are shared between the groups, the mixing weights may differ from one group to another, which allows the group class-conditionals to be different. Summarizing, \cpccInv\ allows group class-conditionals to weigh disjoint regions of $\xset$ differently, thereby allowing a more flexible representation than \ccInv.

\begin{figure}[t]
\centering
\includegraphics[width=\columnwidth]{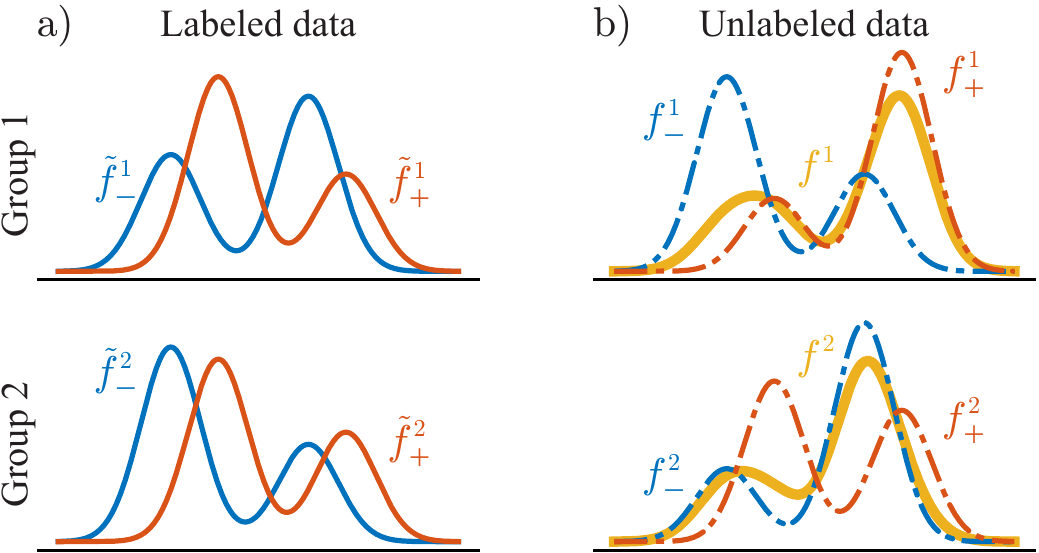} 
\caption{Illustration of the structure and bias considered in this work. \textbf{(a)} Group class-conditionals for the labeled data. $\lbl{f}^g_+(x) = \lbl{p}(x|y=1,g)$ and $\lbl{f}^g_-(x)=\lbl{p}(x|y=0,g)$ are the positive and negative class-conditionals for group $g$, respectively. The solid curves indicate that we observe data from the labeled group class-conditionals. \textbf{(b)} Group class-conditionals and marginals for the unlabeled data. $f^g_+(x) = p(x|y=1,g)$ and $f^g_-(x)=p(x|y=0,g)$ are the positive and negative class-conditionals for group $g$, respectively. $f^g(x) = p(x|g)$ is the group $g$ marginal over $x$. The dashed-dotted curves indicate that we do not observe data from the unlabeled group class-conditionals, but we do observe data from the group marginals. Notice the difference in the class-conditionals across groups and between the labeled and unlabeled data. Also notice that all the positive (negative) group class-conditionals are mixtures sharing the same components, but differing w.r.t. their mixing weights. Additionally, labeled data class priors are generally different from unlabeled data class priors.}
\label{fig1}
\end{figure}

\textbf{Data Assumptions}: Next, consider the following setting for the observed data used to train a binary classifier. Let $\DSU$ be an unlabeled set containing pairs of the form $(x,g) \in \xset \times \gset$ drawn from $p(x,g)$. Let $\DSL$ be a labeled set containing triples of the form $(x,g,y) \in \xset \times \lbl{\gset} \times \yset$, where $\lbl{\gset}$ is a group index set, which might be equal to $\gset$ or a subset/superset of $\gset$ or disjoint from $\gset$. 
Thus, a labeled example may or may not come from the groups that the unlabeled examples belong to. Assume that the labeled triples are drawn from $\lbl{p}\paren*{x,g,y}$, a joint distribution over $\xset \times \lbl{\gset} \times \yset$. That is, we assume that $\lbl{p}\paren*{x,g,y}$ is a biased version of $p(x,g,y)$.  However, similar to $p\paren*{x,g,y}$, $\lbl{p}\paren*{x,g,y}$ follows the \cpccInv\ assumption w.r.t. the same clustering function $\pi$; i.e.,
\begin{align*}
\lbl{p}(x|y,g,\pi(x)) &= \lbl{p}(x|y,\pi(x)).   
\end{align*}
Furthermore, though $\lbl{p}\paren*{x,g,y}$ and $p\paren*{x,g,y}$ are different distributions, we assume them to have equal partition-projected class-conditionals; i.e.,
\begin{align*}
\lbl{p}(x|y,\pi(x)) &= p(x|y,\pi(x)).  
\end{align*}
This assumption is critical for the input-output patterns learned on a labeled data partition to be optimal on the corresponding unlabeled partition, in spite of the bias in the class-conditionals and the group class-conditionals; i.e., $p(x|y) \neq\lbl{p}(x|y)$ and $p(x|y,g) \neq \lbl{p}(x|y,g)$. Figure \ref{fig1} gives an example in which the aforementioned assumptions hold. 

\textbf{Problem Statement}: 
In this paper, we are motivated by the problem of learning a classifier from $\DSL$ and $\DSU$ for optimally predicting class labels for $(x,g)$ drawn from $p(x,g)$. Consider the following two types of classifiers: (1) a group-agnostic classifier that only uses the representation $x$ to separate the positives from the negatives; i.e., a classifier with a score function of the form $s:\xset \rightarrow \extReal$, 
and (2) a group-aware classifier that uses both the representation $x$ and group index $g$; i.e., a classifier with a score function of the form $\group{s}:\xset \times \gset \rightarrow \extReal$. In theory, a group-aware classifier should be at least as good as the group-agnostic one and better than a group-agnostic classifier if the groups carry additional information about the class. In practice, however, a group-aware classifier trained on finite data could be suboptimal  
since more complex models have a higher propensity to overfit. Thus, it is important to theoretically ascertain if the groups carry additional information about the class and only train a group-aware classifier if that is indeed the case. Furthermore, in that case, it is not obvious how to best incorporate the group information when training a group-aware classifier. 

The straightforward approach of encoding the group as a one-hot feature does not explicitly leverage the \cpccInv. Can the groups be incorporated in a manner in which the assumptions are directly leveraged to train a group-aware classifier with improved empirical performance? Additionally, since the labeled data has biased class-conditionals, a classifier trained on the labeled data is likely to be sub-optimal. Can the unlabeled data be exploited by the algorithm to correct for the biases, giving an optimal classification? Our work seeks to answer these questions. 

\section{Contributions}
This paper reveals the following:
\begin{enumerate}
    \item  Knowing the group of an object indeed carries additional information about the class, over that carried by the representation. Further, the improvement in classification brought about by the group can be theoretically quantified in terms of data distribution parameters.
    \item Our novel semi-supervised algorithm is capable of incorporating the groups in a manner that directly exploits the \cpccInv\ assumption to learn a group-aware classifier that is demonstrably better than the group-aware classifier based on one-hot encoding of groups, the group-agnostic classifier, and other ablative models on synthetic and real-world data.
    \item The posterior probability, $p(y=1|x,g)$, can be expressed in terms of quantities that can be estimated from labeled and unlabeled data. 
    \item Despite labeled data bias, our semi-supervised algorithm can learn an optimal classifier under the data assumptions by estimating unbiased distribution parameters on the unlabeled data and using them to correct for the bias.
\end{enumerate}

\section{Theoretical Results}
In this section, we prove that under \cpccInv, the groups indeed carry additional information about the class relative to that carried by the group-agnostic features (Theorem \ref{thm:AUC}). The result is shown in terms of the area under the ROC curve (AUC) improvements brought about by an optimal group-aware classifier, given by the posterior $\group{\rho}(x,g) = p(y=1|x,g)$, over an optimal group-agnostic classifier, given by $\rho(x) = p(y=1|x)$. Additionally, we show that  $\group{\rho}(x,g)$ can be expressed in terms of distribution parameters that can be estimated from labeled and unlabeled data (Theorem \ref{thm:posterior}). The theorem directly informs our algorithm for practical estimation of $\group{\rho}(x,g)$. \par
Let $f_+(x)= p(x|y=1)$ and $f_-(x)= p(x|y=0)$ be the class-conditional densities for the positive and negative class. Let $\group{f}_+(x,g) = p(x,g|y=1) $ and  $\group{f}_-(x,g)=p(x,g|y=0)$ denote the joint density function of the $x$ and $g$ given the class. Theoretically, the AUC of a group-agnostic score function, $s(x)$, can be expressed as 
\begin{align*}
    \AUC(s) &= P\paren*{\tau_s(x_1,x_0) > 1} + \nicefrac{1}{2} P\paren*{\tau_s(x_1,x_0) = 1},
\end{align*}
where $x_1 \sim f_+$, $x_0 \sim f_-$ and $\tau_s(x_1,x_0) = \nicefrac{s(x_1)}{s(x_0)}$. Similarly, the AUC of a group-aware score function, $\group{s}(x,g)$,  can be expressed as 
\begin{align*}
    \AUC(\group{s}) &= P\paren*{\tau_{\group{s}}(x_1, g_1, x_0,g_0) > 1}\\
    & \quad + \nicefrac{1}{2} P\paren*{\tau_{\group{s}}(x_1, g_1, x_0,g_0) = 1},
\end{align*}
where $\tau_{\group{s}}(x_1, g_1, x_0,g_0) = \nicefrac{\group{s}(x_1, g_1)}{\group{s}(x_0, g_0)}$, $(x_1, g_1) \sim \group{f}_+$ and $(x_0, g_0) \sim \group{f}_-$.\par
Next, consider two group-agnostic score functions: (1) posterior-based, $\rho(x) = p(y=1|x)$, and (2) density ratio-based, $r(x) = \nicefrac{f_+(x)}{f_-(x)}$. It can be shown that a density ratio-based score function or any score function that ranks the data points in the same order achieves the highest AUC \cite{uematsu2011theoretically}. Thus $r$ achieves the highest AUC among all score functions defined on $\xset$ and so does $\rho$ since both functions rank the points in $\xset$ in the same order; i.e., $\AUC(r) = \AUC(\rho).$

Further, consider the group-aware posterior as the score function, $\group{\rho}(x,g) = p(y=1|x,g)$. Similar to the discussion above, $\group{\rho}$  achieves the highest AUC  among all score functions defined on $\xset\times\gset$. In Theorem \ref{thm:AUC}, we show that incorporating the group information improves classification as $\AUC(\group{\rho})$ is greater than or equal to $\AUC(\rho)$. The increase in $\AUC(\group{\rho})$ is expressed in terms of random variables $\tau_r^{10} = \nicefrac{r(x_1)}{r(x_0)}$, $\tau_r^{01} = \nicefrac{r(x_0)}{r(x_1)}$, $\omega^{10}=\omega\paren*{g_1,\pi(x_1),g_0,\pi(x_0)}$ and $\omega^{01}=\omega\paren*{g_0,\pi(x_0),g_1,\pi(x_1)}$, where $\omega(g,k,h,j) = \nicefrac{\oddsRatio\paren*{\alpha_{k}^{g}, \alpha_{k}}}{\oddsRatio\paren*{\alpha_{j}^{h}, \alpha_{j}}}$, 
$\alpha_k^g = p(y=1|g,\pi=k)$ is the proportion of positives in partition $k$ for group $g$ and $\alpha_k = p(y=1|\pi = k)$ is the proportion of positives in partition $k$ overall, across all groups; $\oddsRatio(p,q) = \frac{\nicefrac{p}{(1-p)}}{\nicefrac{q}{(1-q)}}$ is the odds ratio between probabilities $p$ and $q$. When $p=q=0$ or $p=q=1$, $\oddsRatio(p,q)$ should be evaluated as $1$.    

Notice that the right-hand side in Theorem \ref{thm:AUC} is non-negative because the indicator and the Dirac delta functions ensure that $\omega^{10}< \tau_r^{01}$. Furthermore, it is strictly positive when $\omega^{10} < \tau_r^{01} \leq 1$ with non-zero probability. Note that $\tau^{10}$ and $\tau^{01}$ depend on feature representations of a pair of positive and negative points, but not on their groups, whereas $\omega^{10}$ and $\omega^{01}$ depend on their groups and the feature representations, however, at a level of granularity, captured by their cluster memberships. 
 Intuitively, the largest increase in AUC is attained when the features in $\xset$ have no predictive power ($f_+ = f_-$); i.e., $\AUC(\rho)=0.5$ and $\forall x\in \xset$, $r(x) = 1$. In this case, $\tau^{10}=1$ with probability $1$. Now, if all group-cluster pairs are pure ($\forall k\in \pset$ and $g\in\gset$, $\alpha_k^g$ is equal to $1$ or $0$), then $\omega^{01}=0$ with probability $1$ and the first and second expectations evaluate to $0.5$ and $0$, respectively, giving $\AUC(\group{\rho})=1$. In general, a large overlap between $f_+$ and $f_-$ gives small values of $\AUC(\rho)$ and thus a more significant increase in $\AUC$ can be expected theoretically, provided the group within a cluster tend to be purer than the cluster. In practice, however, a large overlap would lead to a large variance in estimating group-cluster pair positive proportions (Methods) and, consequently, the expected increase in $\AUC$ of the group-aware classifier might be compromised. In an extreme case, when $f_+=f_-$, the group-cluster pair positive proportions are unidentifiable and cannot be estimated.      
\begin{thm}
\label{thm:AUC} 
(Proof in Appendix) 
Let $\expt$ denote the expectation over $(x_1,g_1){\sim} \group{f}_+$ and $(x_0,g_0) {\sim} \group{f}_-$; $\omega^{10}=\omega(g_1,\pi(x_1), g_0, \pi(x_0))$, $\omega^{01}= \nicefrac{1}{\omega^{10}}$, $\tau_{r}^{10} = \tau_{r}(x_1, x_0)$ and $\tau_{r}^{01} = \nicefrac{1}{\tau_{r}^{10}}$; $\Indtwo{a}{b}{\cdot}$ be the indicator function for the open interval $(a,b)$ and $\dirac{a}{\cdot}$ be the Dirac delta function, evaluating to $1$ at $a\in \real$ and to $0$ otherwise. It follows that

\vspace{-1em}
\begin{align*}
&\AUC(\group{\rho})- \AUC(\rho) = \expt\brac*{\dirac{0}{\omega^{01}}\paren*{\Indtwo{0}{1}{\tau_r^{10}}{+} \nicefrac{1}{2}\dirac{1}{\tau_r^{10}}} }\\ 
&\quad + \expt\brac*{\Indtwo{0}{1}{\omega^{10}}\paren*{\Indtwo{\omega^{10}}{1}{\tau_r^{01}}{ + }\nicefrac{1}{2} \dirac{1}{\tau_r^{01}}}   \paren*{\nicefrac{\tau_r^{01}}{\omega^{10}}{-}1}}.
\end{align*}
\vspace{-1em}
\end{thm}


\begin{thm}
\label{thm:posterior}
(Proof in Appendix) For $\lbl{\rho}(x) = \lbl{p}(y=1|x)$, $\lbl{\alpha}_{k} = \lbl{p}(y=1|\pi =k)$ and $\alpha^g_{k} = p(y=1|g,\pi=k)$,
$$\group{\rho}(x,g) = \paren*{1+ \oddsRatio\paren*{\lbl{\alpha}_{\pi(x)}, \alpha^g_{\pi(x)}}  \frac{1-\lbl{\rho}(x)}{\lbl{\rho}(x)}}^{-1}.$$
\end{thm}

\section{Methods}
In this section, we introduce our main semi-supervised learning algorithm for training an optimal, bias-corrected, group-aware classifier from $\DSL$ and $\DSU$. To this end, we estimate the probability-calibrated  score function $\group{\rho}(x,g)$ by explicitly leveraging \cpccInv\ and the additional assumptions relating the labeled and unlabeled data distribution described in the Problem Formulation section. The algorithm is given by the following steps.
\begin{enumerate}
    \item \textbf{Cluster}: Apply k-means clustering to $\DSL\ \cup\ \DSU$, the combined pool of labeled and unlabeled data to partition $\xset$ into $K$ clusters, $\braces*{\xset_k}_{k=1}^K$. Use the silhouette coefficient \cite{de2015recovering} to determine $K$.
    \item \textbf{Estimate $\bm{\lbl{\rho}(x) = \lbl{p}(y=1|x)}$}: Estimate the labeled data posterior by training a probabilistic classifier on $\DSL$ using group-agnostic features only. Note that a separate classifier may be trained on each labeled cluster to estimate $\lbl{\rho}(x)$, since $\lbl{p}(y=1|x) = \lbl{p}(y=1|x,\pi(x))$.
    \item \textbf{Estimate $\bm{\lbl{\alpha}_{k}=\lbl{p}(y=1|\pi=k)}$}: Estimate the proportion of positives in each cluster in $\DSL$ by counting the positives in the cluster and dividing by the size of the cluster.
    \item \textbf{Estimate $\bm{{\alpha}^g_{k}=p(y=1|g,\pi=k)}$}: Estimate the proportion of positives in each group and cluster pair from $\DSU$ by applying one of the approaches used for domain-adaptation under label shift; see next Section. 
    \item \textbf{Estimate $\bm{\group{\rho}(x,g)=p(y=1|x,g)}$}: Estimate the group-aware posterior by applying the formula derived in Theorem \ref{thm:posterior}, using the estimates of $\lbl{\rho}(x)$, $\lbl{\alpha}_{\pi(x)}$ and $\alpha^g_{\pi(x)}$, computed in the previous steps.
\end{enumerate}


\noindent Except for the estimation of ${\alpha}^g_{k}$ in step 4, all other steps are straightforward to implement. We estimate ${\alpha}^g_{k}$ using techniques from domain-adaptation under label shift as follows. 

\subsection{Estimating Class Proportions under Label Shift}
The key insight for estimation of ${\alpha}^g_{k}$ is that, when restricted to a single cluster, the labeled data and the unlabeled data form an instance of single-source and multi-target domain-adaptation under label shift (same as \ccInv). Let $\mc{L}_k = \braces*{(x,y)| x\in \xset_k, (x,g,y) \in \mc{L}}$ be the subset of $\mc{L}$ containing points from the $k$\textsuperscript{th} cluster only. Let $\mc{U}_k^g = \braces*{x| x\in \xset_k, (x,g) \in \mc{U}}$ be the subset of $\mc{U}$ containing points from the $k$\textsuperscript{th} cluster and $g$\textsuperscript{th} group only. $\mc{L}_k$ serves as the source domain and $\braces*{\mc{U}_k^g}_{g=1}^G$ serves as the $G$ target-domains. Since  \cpccInv\ across groups and the data  assumptions imply $p(x|y,g,\pi(x)) = p(x|y,\pi(x)) = \lbl{p}(x|y,\pi(x))$, the underlying distributions of positives (negatives) in $\mc{L}_k$ and $\mc{U}_k^g$ are equal. The marginal distributions of $x$ corresponding to $\mc{L}_k$ and $\mc{U}_k^g$ only differ in terms of the class proportions. Thus, the label-shift assumptions are satisfied between $\mc{L}_k$ and $\mc{U}_k^g$, which make estimation of ${\alpha}^g_{k}$ feasible. \par
The two state-of-the-art approaches to estimate the target domain class proportions under label-shift are:
\begin{itemize}
    \item Maximum Likelihood Label Shift (MLLS): an Expectation Maximization (EM) based maximum-likelihood estimator of the class proportions that relies on a calibrated probabilistic classifier trained on the source domain \cite{mlls}.
    \item Black Box Shift Estimation (BBSE): a moment-matching based class proportion estimator that works with calibrated or uncalibrated classifier trained on the source domain  \cite{bbse}.  
\end{itemize}
We decided to use MLLS to estimate ${\alpha}^g_{k}$, since it has been shown to outperform BBSE empirically \cite{shrikumar2019adapting}.  Formally, ${\alpha}^g_{k}$ is estimated in an EM framework by iteratively updating it until convergence. Starting with an initial estimate $\hat{\alpha}^g_{k}\ang*{0} = \nicefrac{1}{\abs*{\mc{U}^g_k}} \sum_{x \in \mc{U}^g_k} \lblhat{\rho}(x)$, where $\lblhat{\rho}(x)$ is the group-agnostic posterior estimated from $\mc{L}$ in step 2, it is updated in iteration $t$ as
 $$\hat{\alpha}^g_{k}\ang*{t+1} \leftarrow \frac{1}{\abs*{\mc{U}^g_k}} \sum_{x \in \mc{U}^g_k}\paren*{1+ \oddsRatio\paren*{\lblhat{\alpha}_{k}, \hat{\alpha}^g_{k}\ang*{t}}  \frac{1-\lblhat{\rho}(x)}{\lblhat{\rho}(x)}}^{-1}$$
where $\lblhat{\alpha}_{k}$ is the proportion of positives in $\DSL_k$ computed in step 3. 

\section{Experiments and Empirical Results}
\subsection{Datasets}
The method was evaluated on synthetic and real-world data in two settings: (1) 
\textbf{setting 1}, where the class-conditionals are identical across groups and between labeled and unlabeled data, and (2) 
\textbf{setting 2},  where the class-conditionals vary, but the partition-projected class-conditionals are invariant (\cpccInv\ assumption holds). \par

\textbf{Synthetic Data:} We generated synthetic data from Gaussian mixtures. The positive and negative examples in cluster $k$ were drawn from a pair of $d$-dimensional Gaussian components, N$(\mu^+_{k}, \Sigma^+_{k})$ and N$(\mu^-_{k}, \Sigma^-_{k})$, respectively. Their location and shape parameters were obtained by random perturbations such that the overlap between the pair corresponded to a within-cluster AUC in the range $[0.75, 0.95]$.  
The component pair for each cluster was generated one after the other, further ensuring that they do not overlap significantly with the component pairs already generated. \par
Once the component pairs for all clusters were determined, the data for setting 1 and 2 were generated as follows. Each dataset was generated with 100 groups. The sizes of group $g$ in the labeled ($\abs*{\mc{L}^g}$) and unlabeled ($\abs*{\mc{U}^g}$) data were determined by drawing a random number from N$(1000, 100^2)$ and N$(10000, 1000^2)$, respectively, and rounding to the closest integer. The data dimension ($d$) and the number of clusters ($K$) were picked from  $\{1,2,4,8\}$ and $\{1,4,16,64\}$, respectively, in all pairs of combinations. Next,    
\begin{itemize}
    \item for \textbf{setting 1}, the cluster proportions, $\brac*{\gamma_k}_{k=1}^K$, were sampled from the $K$-dimensional symmetric Dirichlet$(K,2)$. The proportion of positives in cluster $k$ ($\alpha_k$) was sampled from Uniform$(0.01, 0.99)$. Then, $ \gamma_k \alpha_k \abs*{\mc{L}^g}$  and $ \gamma_k \alpha_k \abs*{\mc{U}^g}$ examples (rounded to the closest integer) were sampled from N$(\mu^+_{k}, \Sigma^+_{k})$ as the labeled and unlabeled positives for group $g$, respectively. Similarly, $ \gamma_k (1-\alpha_k) \abs*{\mc{L}^g}$  and $ \gamma_k (1-\alpha_k) \abs*{\mc{U}^g}$ examples were sampled from N$(\mu^-_{k}, \Sigma^-_{k})$ as the labeled and unlabeled negatives for group $g$, respectively.
    \item for \textbf{setting 2}, the group $g$ cluster proportions for the labeled ($\brac*{\lbl{\gamma}_k^g}_{k=1}^K$) and unlabeled ($\brac*{\gamma_k^g}_{k=1}^K$) data were sampled from the $K$-dimensional symmetric Dirichlet$(K,2)$. The proportion of positives in cluster $k$ for group $g$ in labeled ($\lbl{\alpha}_k^g$) and unlabeled ($\alpha_k^g$) data were sampled from Uniform$(0.01, 0.99)$. Then, $ \lbl{\gamma}^g_k \lbl{\alpha}^g_k \abs*{\mc{L}^g}$  and $ \ul{\gamma}^g_k \ul{\alpha}^g_k\abs*{\mc{U}^g}$ examples (rounded to the closest integer) were sampled from N$(\mu^+_{k}, \Sigma^+_{k})$ as the labeled and unlabeled positives for group $g$, respectively. Similarly, $ \lbl{\gamma}^g_k (1-\lbl{\alpha}^g_k) \abs*{\mc{L}^g}$  and $ \ul{\gamma}^g_k (1-\ul{\alpha}^g_k)\abs*{\mc{U}^g}$ examples were sampled from N$(\mu^-_{k}, \Sigma^-_{k})$ as the labeled and unlabeled negatives for group $g$, respectively
\end{itemize}

\noindent \textbf{Real-World Data:} Three binary classification datasets, generated from the Folktables American Community Survey (ACS) data \cite{folktables}, were used for evaluation: Income, Income Poverty Ratio (IPR), and Employment (Table \ref{tab:dataset}, Appendix). Additional high-dimensional embeddings data can also be found in the Appendix, available on arXiv.

 For each dataset, labeled ($\mc{L}$) and unlabeled ($\mc{U}$) samples for the two settings were generated from the pool of available labeled examples $\mc{D}$. State was used as the group variable in the ACS datasets. The sets of all group $g$ examples in $\mc{D}$ were first equally divided into labeled ($L^g$) and unlabeled ($U^g$) pools. The final labeled ($\mc{L}^g$) and unlabeled ($\mc{U}^g$) sets for group $g$ were generated by resampling with replacement from $L^g$ and $U^g$. 
To this end, $\mc{D}$ was first partitioned into $K$ clusters by running mini-batch k-means with batch size 4096, where $K$ was chosen from $\{1,2,4,8\}$, to maximize the silhouette coefficient on a random sample of 25,000 examples. \par
To create a setting 1 dataset, first, $\brac*{\gamma_k}_{k=1}^K$ and $\alpha_k$ were sampled similarly as for the synthetic datasets. Then, $\gamma_k \alpha_k |L^g|$  and $\gamma_k \alpha_k |U^g|$ examples sampled from the positives in $L^g$ and $U^g$, lying in cluster $k$, were added to $\mc{L}^g$ and $\mc{U}^g$, respectively. Similarly, $\gamma_k (1-\alpha_k) |L^g|$  and $\gamma_k (1-\alpha_k) |U^g|$ examples sampled from the negatives in $L^g$ and $U^g$, lying in cluster $k$, were also added to $\mc{L}^g$ and $\mc{U}^g$.

To create a setting 2 dataset, first, $\brac*{\lbl{\gamma}^g_k}_{k=1}^K$, $\brac*{\gamma^g_k}_{k=1}^K$, $\lbl{\alpha}^g_k$ $\alpha^g_k$ were sampled similarly as for the synthetic datasets. Then  $\lbl{\gamma}^g_k \lbl{\alpha}^g_k |L^g|$  and $\gamma^g_k \alpha^g_k |U^g|$ examples sampled from the positives in $L^g$ and $U^g$, lying in cluster $k$, were added to $\mc{L}^g$ and $\mc{U}^g$, respectively. Similarly, $\lbl{\gamma}^g_k (1-\lbl{\alpha}^g_k) |L^g|$  and $\gamma^g_k (1-\alpha^g_k) |U^g|$ examples sampled from the negatives in $L^g$ and $U^g$, lying in cluster $k$, were also added to $\mc{L}^g$ and $\mc{U}^g$.

\subsection{Experimental Protocol}
Experiments were run at least 10 times for each dataset, repeating the data generation process each time. The performance of each method was measured using AUC on a held-out set of 20\% of the groups of unique examples in $\mathcal{U}$, averaged over all repetitions.

A held-out validation set was constructed by randomly removing 20\% of groups of unique examples in $\mathcal{L}$. Our method fits a mini-batch k-means model to $\mathcal{L} \cup \mathcal{U}$ with batch size 4096, to estimate the clustering used to generate the data. $K \in \{1,2,4,8\}$ is selected to maximize the silhouette coefficient estimated on a random batch of 25,000 examples. A random forest of 500 decision trees with maximum depth of 10 was fit to each cluster in the labeled training data, splitting on the gini criterion. All classifiers were calibrated by Platt's scaling \cite{Platt1999} using a held-out validation set. The MLLS algorithm was run for 100 iterations to estimate the unlabeled group-cluster class priors ${\alpha}^g_{k}$.

The method was compared to five baseline methods: (1) Global, where a single classifier is trained on the labeled examples, (2) Group-Aware Global, where the group-agnostic feature representation of each example is concatenated with a one-hot encoded vector of that example's group, (3) Cluster Global, where a separate classifier is trained on each cluster in the labeled data, (4) Label Shift, where a single classifier is trained on the labeled examples and then adjusted to the estimated class prior of each group's unlabeled distribution, and (5) CORAL, a domain adaptation method where a separate classifier is trained for each group, aligning the second order statistics of the labeled examples to those of the group's unlabeled examples \cite{Sun2016}. Note that Label Shift is a special case of our method ($K=1$). The same model training, selection, and calibration procedures, as described above, were used in all baseline methods.

\subsection{Comparative Experiments on Synthetic Data}
\begin{figure}[t]
\centering
\includegraphics[width=0.8\columnwidth]{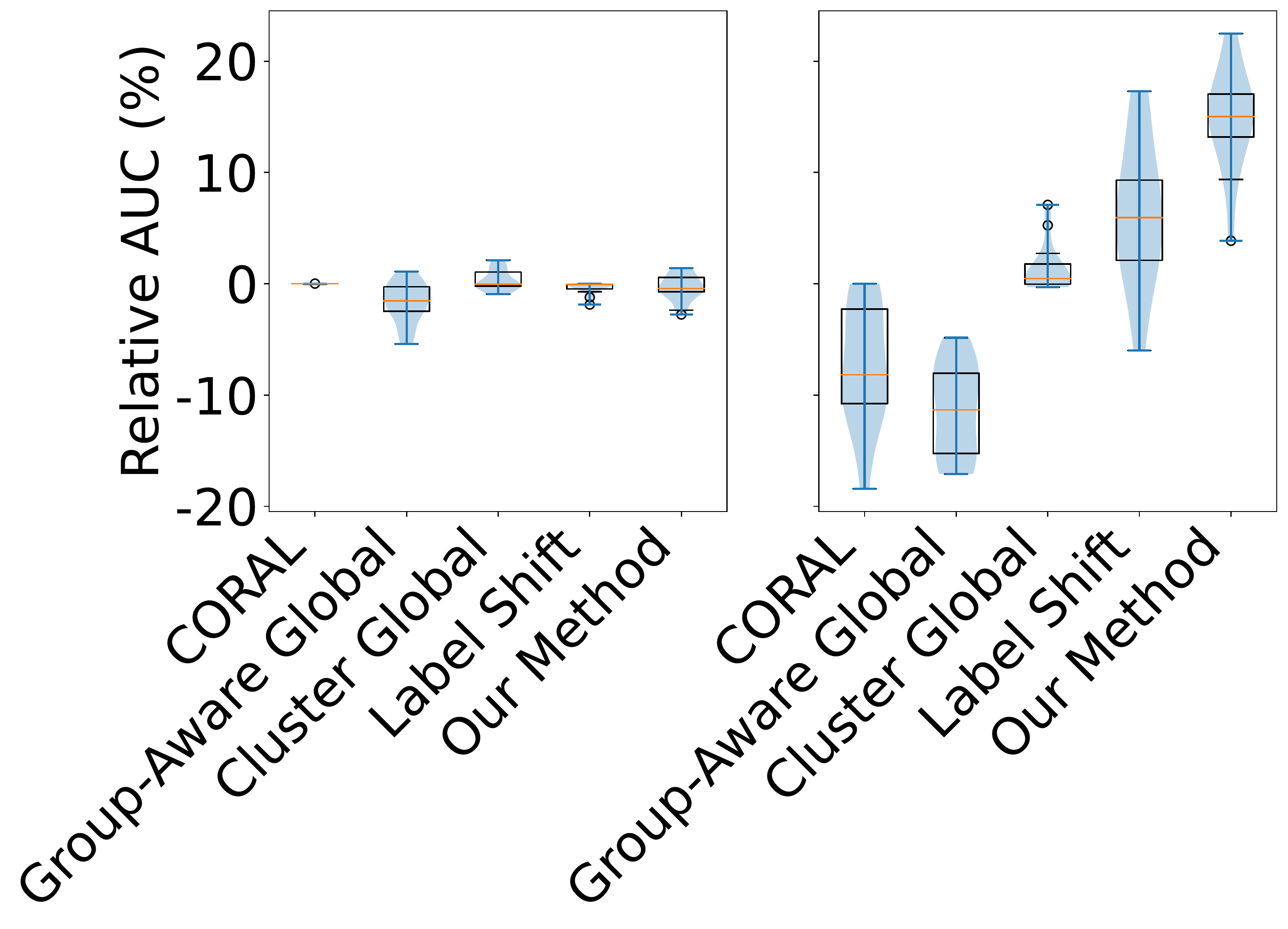}
\caption{Distribution of average AUCs, calculated on the held-out test set, relative to the AUC of Global on synthetic datasets. Left: setting 1, Right: setting 2.}.
\label{synthfig}
\end{figure}

\begin{figure}[t]
\centering
\includegraphics[width=\columnwidth]{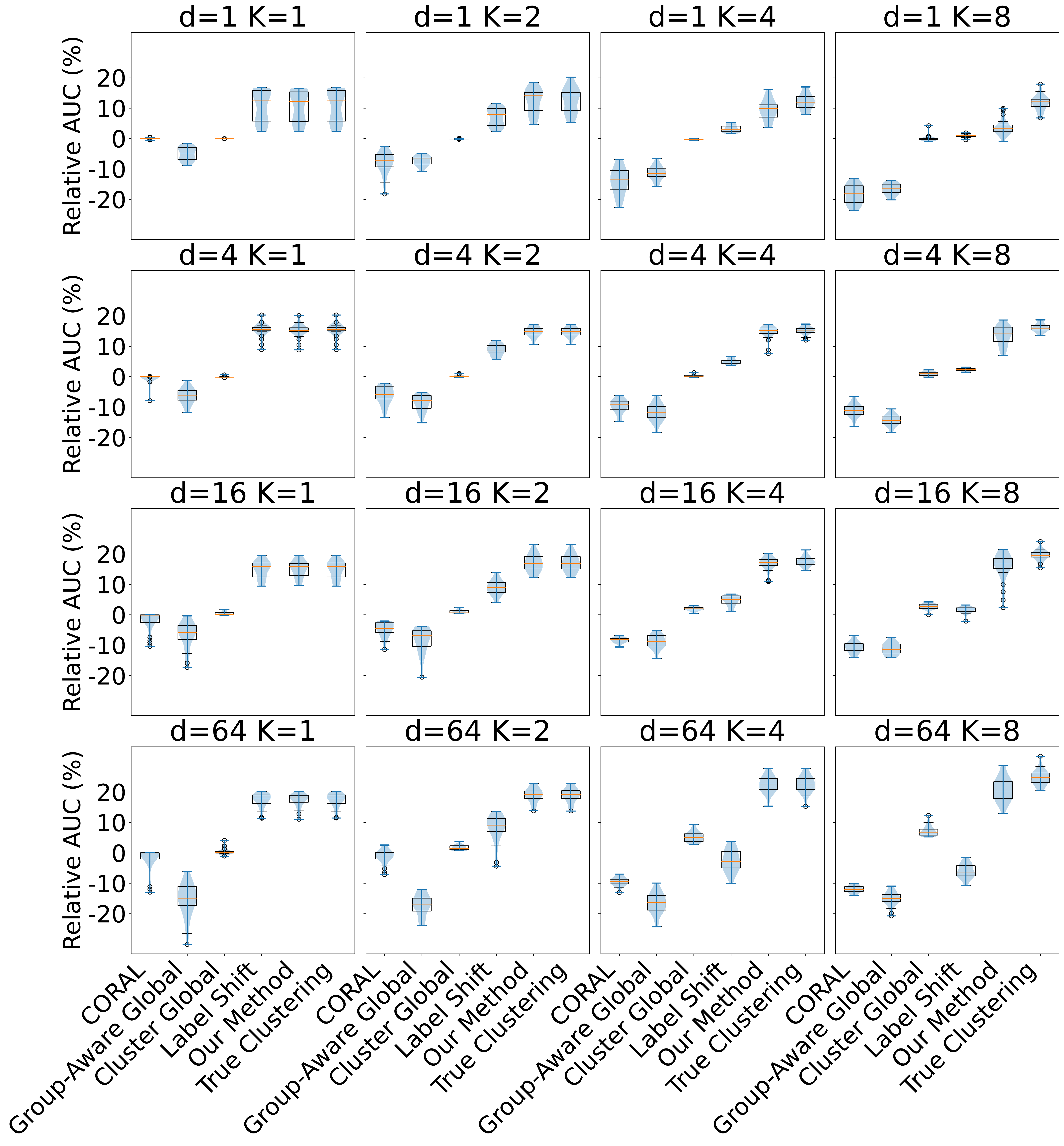}
\caption{Distribution of AUCs, calculated on the held-out test set, relative to the AUC of Global on synthetic datasets in setting 2 separated by $d$ and $K$.}
\label{synthfigbreakdown2}
\end{figure}

Figure \ref{synthfig} shows the distribution of AUCs calculated on the test data relative to the global classifier, with the AUCs for each dimensionality-cluster pair averaged over all iterations. The left sub-figure shows that our method maintains performance even when applied on datasets in which examples are all drawn from the same distribution. The right sub-figure shows that our method leads to substantial improvement in AUC 
when examples are drawn from distributions with the assumed bias model. Figure \ref{synthfigbreakdown2} shows the effects of $d$ and $K$ on the relative performance of our method. Our method maintains strong performance when applied to datasets with both high dimensionality and many clusters.

\subsection{Comparative Experiments on Real-World Data}
Tables \ref{tab:realSetting1} and \ref{tab:realSetting2} list the AUCs calculated on the test set, averaged over all repetitions, for real-world datasets generated in settings 1 and 2, respectively. Table \ref{tab:realSetting1} shows that our method maintains comparable performance to that of Global despite the data lacking the bias our method aims to address. The experiments summarized in Table \ref{tab:realSetting2} demonstrate the theoretical performance improvements our method can realize when presented with real-world biased data.

\begin{table}
\begin{tabular}{lrrr}
Method & Income & Employment &   IPR \\
\hline
CORAL              &  0.865 &      0.883 & 0.792 \\
Global             &  0.879 &      0.903 & 0.825 \\
Group-Aware Global &  0.870 &      0.897 & 0.812 \\
Cluster Global     &  0.879 &      0.903 & 0.828 \\
Label Shift        &  0.873 &      0.900 & 0.812 \\
Our Method         &  0.874 &      0.898 & 0.812 \\
\hline
True Clustering               &  0.873 &      0.900 & 0.812 \\
\end{tabular}
\caption{Average AUC calculated on the held-out test set for real-world datasets in setting 1.}
\label{tab:realSetting1}
\end{table}

\begin{table}
\begin{tabular}{lrrr}
Method & Income & Employment &   IPR \\
\hline
CORAL              &  0.853 &      0.851 & 0.760 \\
Global             &  0.870 &      0.897 & 0.816 \\
Group-Aware Global &  0.846 &      0.865 & 0.768 \\
Cluster Global     &  0.871 &      0.898 & 0.820 \\
Label Shift        &  0.876 &      0.904 & 0.831 \\
Our Method         &  0.884 &      0.916 & 0.850 \\
\hline
True Clustering               &  0.900 &      0.927 & 0.862 \\
\end{tabular}
\caption{Average AUC calculated on the held-out test set for real-world datasets in setting 2.}
\label{tab:realSetting2}
\end{table}

\subsection{Effects of Clustering}
Because our method fits a clustering model on the resampled dataset, not the original dataset that is used to find the clustering during data generation, it is possible that our method does not recover the true clustering used to generate the observed samples. These estimated and true clusterings can differ in the cluster centers and number of clusters. To analyze the effect the quality of the clustering has on classification performance, we compare our method's performance to that of our method when given access to the true clustering used to generate the data. Figure \ref{synthfigbreakdown2} shows the extent to which the performance of our method is affected by estimating the clustering on resampled data in the experiments. The performance degrades with the increasing number of clusters, becoming noticeable at $K=8$. Table \ref{tab:realSetting2} shows that additional improvements in performance could be made with access to the true clustering of the data in the case of real-world datasets with bias.

\section{Related Work}
Since the early work by \citeauthor{heckman1979sample} (\citeyear{heckman1979sample}), the problem of learning from biased data has been extensively studied under sample selection bias, dataset shift, transfer learning and domain adaptation \cite{Peng2003, zadrozny2004learning, Dudik2005, Ben-David2006, Huang2006, cortes2008sample, storkey2009training, daume2009frustratingly, Pan2010,Hsieh2019, Jain2020, labelShiftUnified, dePaolis2023}. Most recently, learning predictive models in settings where the training and test distributions differ has been studied extensively in the literature of domain adaptation \cite{Kouw2019, labelShiftUnified}. Our problem relates to this literature in that it can be reformulated as an instance of multi-target unsupervised domain adaptation by considering the collection of all labeled examples as the source domain and the unlabeled examples from each group as a distinct target domain with a novel assumption placed on the distributions of these domains. 

Several methods, including MLLS \cite{mlls} and BBSE \cite{bbse}, have been developed to correct for label shift, which we refer to as \ccInv, for single-target unsupervised domain adaptation and have been shown to relate to the  distribution-matching methods \cite{labelShiftUnified}. While not explored in these works, under the CC-invariance assumption only, it would be straightforward to apply these distribution matching approaches in isolation to each group to correct for differences in class priors. In a less rigid PCC-invariance assumption shared across the labeled and unlabeled groups, it is assumed that both prior and class-conditional distributions differ across groups, yet the bias correction remains practical by generalizing the methodology developed to address CC-invariance.


While we are unaware of any previous work proposing the PCC-invariance assumption, there exists prior work on covariate shift that assumes different class-conditional distributions across groups. In the context of regression, flexibility has been given to class-conditional distributions across training and test sets by introducing a latent variable $r$ and assuming $p_{\textrm{train}}(x,y|r) = p_{\textrm{test}}(x,y|r)$ and $p_{\textrm{train}}(r) \neq p_{\textrm{test}}(r)$ \cite{mixtureRegression}.

\section{Conclusions}
The partition-projected class-conditional invariance (\cpccInv) assumption, introduced in this paper, is a flexible means of modeling bias and structure in binary classification problems with grouped data. In contrast to the existing bias models, it allows both the posterior ($p(y|x)$) and the class-conditional ($p(x|y)$) in the labeled data to be biased and yet facilitates theoretically sound bias correction with the aid of unbiased unlabeled data. 
Moreover, as a model to capture structure between groups, under \cpccInv\, a group-aware classifier achieves a provably improved AUC over a group-agnostic classifier. \par
By directly exploiting \cpccInv, our semi-supervised algorithm is an effective approach for bias correction and exploiting the group structure present in the data, as demonstrated by improved performance over group-agnostic and group-aware classifiers on real and synthetic data. Furthermore, the approach is robust to the absence of bias and structure in the data as demonstrated by the experiments. Lastly, the approach has a desirable property of learning a probability-calibrated classifier. 

\section*{Acknowledgements}
Clara De Paolis Kaluza for proofreading and NIH awards U01 HG012022 and R01 HD101246. Code is available at https://github.com/Dzeiberg/leveraging\_structure.

\bibliography{aaai23}

\appendix
\onecolumn
\setcounter{thm}{0}
\section{Theorems and Proofs}

\begin{thm}
Let $p(x,g,y)$ be a joint distribution on $\xset\times\gset\times\yset$ satisfying \cpccInv\ w.r.t. the partitioning function $\pi:\xset \rightarrow \pset$ as described in Problem Formulation.
For $\rho(x) = p(y=1|x)$ and $\group{\rho}(x,g) = p(y=1|x,g)$,


\begin{align*}
\AUC(\group{\rho})&= \AUC\paren*{\rho} + \expt\brac*{\indc{(0,1)}{\omega^{10}}\paren*{\indc{\paren*{\omega^{10}, 1}}{\tau_r^{01}} + \frac{1}{2} \dirac{1}{\tau_r^{01} }}\paren*{\frac{\tau_r^{01}}{\omega^{10}} -1}} + \expt\brac*{\paren*{\Ind{\paren*{0, 1}}{\tau_r^{10}}+ \frac{1}{2}\dirac{1}{\tau_r^{10}}} \times \dirac{0}{\omega^{01}}}
\end{align*}

\noindent where the expectation is taken over $(x_1,g_1){\sim} \group{f}_+$ and $(x_0,g_0) {\sim} \group{f}_-$; 
$\omega^{10}=\omega(g_1,\pi(x_1), g_0, \pi(x_0))$ and $\omega^{01}=\nicefrac{1}{\omega^{10}}$; $\tau_{r}^{10} = \tau_{r}(x_1, x_0)$ and $\tau_{r}^{01} = \nicefrac{1}{\tau_{r}^{10}}$; $\indc{A}{\cdot}$ is the indicator function for set $A$ and $\dirac{a}{\cdot}$ be the Dirac delta function, evaluating to $1$ at $a\in \real$ and to $0$ otherwise. 

\begin{proof}
First we derive the following expression for the group-aware posterior $\group{\rho}(x,g)=p(y=1|x,g)$ can be expressed as
\begin{align}
\group{\rho}(x,g)
& = p\paren*{y=1|x,g,\pi(x)} \notag\\
& = \frac{p\paren*{y=1|g,\pi(x)}p\paren*{x|y=1,g,\pi(x)}}{p\paren*{x|g,\pi(x)}} \notag\\
& = \frac{p\paren*{y=1|g,\pi(x)}p\paren*{x|y=1,\pi(x)}}{p\paren*{x|g,\pi(x)}} \tag{\cpccInv\ across groups} \notag\\
& = \frac{p\paren*{y=1|g,\pi(x)}p\paren*{x|y=1,\pi(x)}}{p\paren*{y=1|g,\pi(x)}p\paren*{x|y=1,\pi(x)}+p\paren*{y=0|g,\pi(x)}p\paren*{x|y=0,\pi(x)}} \tag{\cpccInv\ across groups}\notag \\
& = \frac{\alpha^g_{\pi(x)} \cl{f}_+(x)}{\alpha^g_{\pi(x)} \cl{f}_+(x)+\paren*{1-\alpha^g_{\pi(x)}} \cl{f}_-(x)}, \label{eq:postgrp}
\end{align}
where $\alpha^g_{k}=p(y=1|g, \pi=k)$ is the proportion of positives in group $g$ examples coming from partition $k$;  $\cl{f}_+(x)= p(x|y=1, \pi(x))$ and $\cl{f}_-(x)= p(x|y=0, \pi(x))$ are the class-conditionals projected on the partition containing $x$.  
Next, we derive the following expression for $\cl{f}_+(x)$. 
\begin{align}
    \cl{f}_+(x) &= p(x|y=1,\pi(x))\notag\\
    & = \frac{p(x, \pi(x)|y=1)}{p(\pi(x)|y=1)}\notag\\
    & = \frac{f_+(x)}{\frac{p(\pi(x))p(y=1|\pi(x))}{p(y=1)}}\notag\\
    &=\frac{\alpha f_+(x)}{\alpha_{\pi(x)}\gamma_{\pi(x)}}, \label{eq:clustercc1}
\end{align}
where $\alpha =p(y=1)$, the proportion of positives in the entire distribution; 
$\alpha_k =p(y=1|\pi=k)$, the proportion of positives in the partition $k$;  $\gamma_{k}=p\paren*{\pi=k}$, the proportion of examples lying in partition $k$ from the entire distribution. Note that if $f_+(x)=0$ on the entire partition $\pi(x)$, containing point $x$, $\alpha_{\pi(x)}$ is equal to 0 as well, which leads to $0/0$ on the right hand side. Since $f'_+(x)=0$ on the entire partition $\pi(x)$ in this case, we consider the RHS to be equal to 0. 
Similarly,
\begin{align}
    \cl{f}_-(x) &=\frac{(1-\alpha) f_-(x)}{\paren*{1-\alpha_{\pi(x)}}\gamma_{\pi(x)}}. \label{eq:clustercc2}
\end{align}
Again note that if $f_-(x)=0$ on the entire partition $\pi(x)$, containing point $x$, $1-\alpha_{\pi(x)}$ is equal to 0 as well, which leads to $0/0$ on the right hand side. Since $f'_-(x)=0$ on the entire partition $\pi(x)$ in this case, we consider the RHS to be equal to 0. 

\noindent Next, we derive the following expression from Eq. \ref{eq:postgrp}.
\begin{align}
    \frac{\group{\rho}(x,g)}{1-\group{\rho}(x,g)} &= \frac{\alpha^{g}_{\pi(x)} \cl{f}_+(x)}{\paren*{1-\alpha^{g}_{\pi(x)}} \cl{f}_-(x)}\notag\\
    &=\frac{\frac{\alpha\alpha^{g}_{\pi(x)}}{\alpha_{\pi(x)}\gamma_{\pi(x)}}f_+(x)}{\frac{(1-\alpha)\paren*{1-\alpha^{g}_{\pi(x)}}}{\paren*{1-\alpha_{\pi(x)}}\gamma_{\pi(x)}}f_-(x)} \tag{from Eq. \ref{eq:clustercc1} and \ref{eq:clustercc2}} \\
     &=\frac{\alpha}{1-\alpha}\oddsRatio\paren*{\alpha^{g}_{\pi(x)},\alpha_{\pi(x)}}r(x), \label{eq:postGrpRatio}
\end{align}
where $r(x)=\nicefrac{f_+(x)}{f_-(x)}$; $\oddsRatio(p,q)=\frac{\nicefrac{p}{(1-p)}}{\nicefrac{q}{(1-q)}}$ is the odds ratio between probabilities $p$ and $q$. Defined as such, $\oddsRatio(p,q)$ is not well-defined when $p=q=0$ or $p=q=1$. We extend the definition in the two cases as $\oddsRatio(p,q)=1$. Due to this extension, Eq. \ref{eq:postgrp} still holds true when $\alpha_{\pi(x)}$ is equal to $0$ or $1$. Precisely, $\alpha_{\pi(x)}=0$ means that partition $\pi(x)$ does not contain any positives and consequently $r(x)=0$, $\group{\rho}(x,g)=0$ and $\alpha^g_{\pi(x)}=0$ for any $g\in\gset$. Thus both RHS and LHS are equal. Similarly, $\alpha_{\pi(x)}=1$ means that partition $\pi(x)$ does not contain any negatives and consequently $r(x)=\infty$, $\group{\rho}(x,g)=1$ and $\alpha^g_{\pi(x)}=1$ for any $g\in\gset$. Thus both RHS and LHS are equal. 

\noindent Next, for inputs $x_0, x_1 \in \xset$ and groups $g_0, g_1 \in \gset$,
\begin{align*}
    \group{\rho}(x_1,g_1) > \group{\rho}(x_0,g_0) & \Leftrightarrow  \frac{ \group{\rho}(x_1,g_1)}{1- \group{\rho}(x_1,g_1)} > \frac{ \group{\rho}(x_0,g_0)}{1- \group{\rho}(x_0,g_0)}\\
   & \Leftrightarrow \oddsRatio\paren*{\alpha^{g_1}_{\pi(x_1)},\alpha_{\pi(x_1)}}r(x_1) > \oddsRatio\paren*{\alpha^{g_0}_{\pi(x_0)},\alpha_{\pi(x_0)}}r(x_0) \\
    & \Leftrightarrow  \tau_r(x_1, x_0) > \omega\paren*{g_0, \pi(x_0), g_1, \pi(x_1)},
\end{align*}
where $\tau_r(x_1,x_0) = \nicefrac{r(x_1)}{r(x_0)}$ and  $\omega(g,k,h,j) = \frac{\oddsRatio\paren*{\alpha_{k}^{g}, \alpha_{k}}}{\oddsRatio\paren*{\alpha_{j}^{h}, \alpha_{j}}}$.   Similarly,
\begin{align*}
    \group{\rho}(x_1,g_1) = \group{\rho}(x_0,g_0)
    & \Leftrightarrow   \tau_r(x_1, x_0) = \omega\paren*{g_0, \pi(x_0), g_1, \pi(x_1)}.
\end{align*}
\noindent Note that for a positive $x \in \xset$ belonging to group $g\in \gset$, $\alpha^{g}_{\pi(x)} \neq 0$ and $\alpha_{\pi(x)} \neq 0$. Thus $\oddsRatio\paren*{\alpha^{g}_{\pi(x)},\alpha_{\pi(x)}} \neq 0$. Similarly for a negative $x \in \xset$ belonging to group $g\in \gset$ $\alpha^{g}_{\pi(x)} \neq 1$ and $\alpha_{\pi(x)} \neq 1$. Thus $\oddsRatio\paren*{\alpha^{g}_{\pi(x)},\alpha_{\pi(x)}} \neq \infty$. Now, if $x_1$ is positive and $x_0$ is negative, $\omega\paren*{g_0, \pi(x_0), g_1, \pi(x_1)}$ is well-defined, finite and could potentially take $0$ value. In contrast, if $x_0$ is positive and $x_1$ is negative, $\omega\paren*{g_0, \pi(x_0), g_1, \pi(x_1)}$ is well-defined and non zero, but not necessarily finite. \par
\noindent Let $P_{10}$ and $\expt_{10}$ denote the probabilities and expectations computed with $(x_1,g_1) \sim \bar{f}_+(x,g)= p(x,g|y=1)$ and $(x_0,g_0) \sim \bar{f}_-(x,g)= p(x,g|y=0)$. Similarly, $P_{01}$ and $\expt_{01}$  denote the probabilities and expectations computed with $(x_0,g_0) \sim \bar{f}_+(x,g)$ and $(x_1,g_1) \sim \bar{f}_-(x,g)$. Further, let $\omega^{10} = \omega(g_1,\pi(x_1), g_0, \pi(x_0))$, $\omega^{01} = \omega(g_0,\pi(x_0), g_1, \pi(x_1))$, $\tau_r^{10} = \tau_r(x_1,x_0)$ and $\tau_r^{01} = \tau_r(x_0,x_1)$. Note that $\omega^{01} = \nicefrac{1}{\omega^{10}}$ and  $\tau_r^{01} = \nicefrac{1}{\tau_r^{10}}$ Now,

\begin{align*}
    \AUC\paren*{\group{\rho}} &= P_{10}\paren*{\group{\rho}(x_1, b_1) > \group{\rho}(x_0, b_0)} + \frac{1}{2} P_{10}\paren*{\group{\rho}(x_1, b_1) = \group{\rho}(x_0, b_0)}  \notag\\
    &= P_{10}\paren*{\tau_r^{10} > \omega^{01}}  + \frac{1}{2} P_{10}\paren*{\tau_r^{10} = \omega^{01}} \notag\\
    &= P_{10}\paren*{\tau_r^{10} > \omega^{01} \andd \omega^{01}\geq 1}  + P_{10}\paren*{\tau_r^{10} > \omega^{01} \andd \omega^{01}< 1} + \frac{1}{2} P_{10}\paren*{\tau_r^{10} = \omega^{01}} \notag\\
    &= P_{10}\paren*{\tau_r^{10} > 1 \andd \omega^{01}\geq 1} - P_{10}\paren*{1 < \tau_r^{10} \leq \omega^{01} \andd \omega^{01}> 1}  \notag\\
    & \quad + P_{10}\paren*{\tau_r^{10} >1 \andd \omega^{01}< 1}  + P_{10}\paren*{1 \geq \tau_r^{10} > \omega^{01} \andd \omega^{01}< 1}+ \frac{1}{2} P_{10}\paren*{\tau_r^{10} = \omega^{01}} \notag\\
    &=P_{10}\paren*{\tau_r^{10} > 1} +P_{10}\paren*{1 \geq \tau_r^{10} > \omega^{01}\ \&\ \omega^{01} < 1} - P_{10}\paren*{1 < \tau_r^{10} \leq \omega^{01}\ \&\ \omega^{01} > 1}  \notag\\
    & \quad +\frac{1}{2} P_{10}\paren*{\tau_r^{10} = 1}+\frac{1}{2} P_{10}\paren*{\tau_r^{10} = \omega^{01}}  -\frac{1}{2} P_{10}\paren*{\tau_r^{10} = 1} \notag\\
    &= \AUC\paren*{\rho}  + P_{10}\paren*{1 > \tau_r^{10} > \omega^{01}\ \&\ \omega^{01} < 1} + P_{10}\paren*{\tau_r^{10} = 1\ \&\ \omega^{01} < 1} \notag\\
    & \quad - P_{10}\paren*{1 < \tau_r^{10} < \omega^{01}\ \&\ \omega^{01} > 1} -P_{10}\paren*{\tau_r^{10} = \omega^{01}\ \&\ \omega^{01} > 1} \tag{because AUC($\rho$) = AUC($\tau_r$)}\\
    &\quad +\frac{1}{2} P_{10}\paren*{\tau_r^{10} = \omega^{01}\ \&\ \omega^{01} \neq 1}  -\frac{1}{2} P_{10}\paren*{\tau_r^{10} = 1 \ \&\ \omega^{01} \neq 1} \notag\\
    &= \AUC\paren*{\rho} +P_{10}\paren*{1 > \tau_r^{10} > \omega^{01}\ \&\ \omega^{01} < 1} - P_{10}\paren*{1 < \tau_r^{10} < \omega^{01}\ \&\ \omega^{01} > 1}\notag\\
    & \quad +\frac{1}{2} \paren*{P_{10}\paren*{\tau_r^{10} = \omega^{01} \ \&\ \omega^{01} < 1} -P_{10}\paren*{\tau_r^{10} = \omega^{01}\ \&\ \omega^{01} > 1}} 
    \notag\\
    & \quad +\frac{1}{2} \paren*{P_{10}\paren*{\tau_r^{10} = 1\ \&\ \omega^{01} < 1}   - P_{10}\paren*{\tau_r^{10} = 1 \ \&\ \omega^{01} > 1}} \notag\\
\end{align*}
As demonstrated earlier, since $x_1$ is positive and $x_0$ is negative under $P_{10}$, $\omega^{01}$ is well-defined and finite. Furthermore it is non-negative by definition. In other words $0\leq\omega^{01}<\infty$ with probability $1$ under $P_{10}$. Thus
\begin{align*}
    \AUC\paren*{\group{\rho}} &= \AUC\paren*{\rho} +P_{10}\paren*{1 > \tau_r^{10} > \omega^{01}\ \&\ 0\leq \omega^{01} < 1} - P_{10}\paren*{1 < \tau_r^{10} < \omega^{01}\ \&\ \infty >\omega^{01} > 1}\notag\\
    & \quad +\frac{1}{2} \paren*{P_{10}\paren*{\tau_r^{10} = \omega^{01} \ \&\ 0 \leq \omega^{01} < 1} -P_{10}\paren*{\tau_r^{10} = \omega^{01}\ \&\ \infty >\omega^{01} > 1}} 
    \notag\\
    & \quad +\frac{1}{2} \paren*{P_{10}\paren*{\tau_r^{10} = 1\ \&\ 0\leq \omega^{01} < 1}   - P_{10}\paren*{\tau_r^{10} = 1 \ \&\ \infty >\omega^{01} > 1}} \notag\\
    &= \AUC\paren*{\rho} +P_{10}\paren*{1 > \tau_r^{10} > \omega^{01}\ \&\ 0< \omega^{01} < 1} - P_{10}\paren*{1 < \tau_r^{10} < \omega^{01}\ \&\ \infty >\omega^{01} > 1}\notag\\
    & \quad +\frac{1}{2} \paren*{P_{10}\paren*{\tau_r^{10} = \omega^{01} \ \&\ 0 < \omega^{01} < 1} -P_{10}\paren*{\tau_r^{10} = \omega^{01}\ \&\ \infty >\omega^{01} > 1}} 
    \notag\\
    & \quad +\frac{1}{2} \paren*{P_{10}\paren*{\tau_r^{10} = 1\ \&\ 0< \omega^{01} < 1}   - P_{10}\paren*{\tau_r^{10} = 1 \ \&\ \infty >\omega^{01} > 1}} \notag\\
    & \quad + P_{10}\paren*{1 > \tau_r^{10} > 0 \ \&\ \omega^{01} = 0} + \frac{1}{2} P_{10}\paren*{\tau_r^{10} = 0 \ \&\  \omega^{01} = 0}  +\frac{1}{2} P_{10}\paren*{\tau_r^{10} = 1\ \&\ \omega^{01} = 0}\\
     &= \AUC\paren*{\rho} +P_{10}\paren*{1 > \tau_r^{10} > \omega^{01}\ \&\ 0< \omega^{01} < 1} - P_{10}\paren*{1 < \tau_r^{10} < \omega^{01}\ \&\ \infty >\omega^{01} > 1}\notag\\
    & \quad +\frac{1}{2} \paren*{P_{10}\paren*{\tau_r^{10} = \omega^{01} \ \&\ 0 < \omega^{01} < 1} -P_{10}\paren*{\tau_r^{10} = \omega^{01}\ \&\ \infty >\omega^{01} > 1}} 
    \notag\\
    & \quad +\frac{1}{2} \paren*{P_{10}\paren*{\tau_r^{10} = 1\ \&\ 0< \omega^{01} < 1}   - P_{10}\paren*{\tau_r^{10} = 1 \ \&\ \infty >\omega^{01} > 1}} \notag\\
    & \quad + P_{10}\paren*{1 > \tau_r^{10} > 0 \ \&\ \omega^{01} = 0}  +\frac{1}{2} P_{10}\paren*{\tau_r^{10} = 1\ \&\ \omega^{01} = 0} \tag{because $\tau_r^{10} > 0$ with probability 1 under $P_{10}$}
\end{align*}

\noindent Now converting probabilities to expectations,
\begin{align}
    \AUC\paren*{\group{\rho}} &= \AUC\paren*{\rho} + \expt_{10}\brac*{\Ind{\paren*{\omega^{01}, 1}}{\tau_r^{10}} \times\Ind{(0,1)}{\omega^{01}}} - \expt_{10}\brac*{\Ind{\paren*{1,\omega^{01}}}{\tau_r^{10}} \times \Ind{(1,\infty)}{\omega^{01}}}\notag\\
    & \quad +\frac{1}{2} \paren*{\expt_{10}\brac*{\dirac{ \omega^{01}}{\tau_r^{10}} \times\Ind{(0,1)}{\omega^{01}}} -\expt_{10}\brac*{\dirac{ \omega^{01}}{\tau_r^{10}} \times\Ind{(1,\infty)}{\omega^{01}}}} 
    \notag\\
    & \quad +\frac{1}{2} \paren*{\expt_{10}\brac*{\dirac{1}{\tau_r^{10} }\times\Ind{(0,1)}{\omega^{01}}}   - \expt_{10}\brac*{\dirac{1}{\tau_r^{10} }\times\Ind{(1,\infty)}{\omega^{01}}}} \notag \\
    &\quad + \expt_{10}\brac*{\Ind{\paren*{0, 1}}{\tau_r^{10}}\times \dirac{0}{\omega^{01}}}  +\frac{1}{2} \expt_{10}\paren*{\dirac{1}{\tau_r^{10}}\times \dirac{0}{\omega^{01}}} \label{eq:expt}
\end{align}
To simplify the expressions further, we first derive the following expression

\begin{align}
    \frac{\group{f}_+(x_1,g_1)\group{f}_-(x_0,g_0)}{\group{f}_+(x_0,g_0)\group{f}_-(x_1,g_1)} &= \frac{p(x_1,g_1|y_1 = 1) p(x_0,g_0|y_0 = 0)}{p(x_1,g_1|y_1 = 0) p(x_0,g_0|y_0 = 1)} \notag \\
    &= \frac{\frac{p(y_1 = 1|x_1,g_1)}{p(y_1=1)} \frac{p(y_0 = 0|x_0,g_0)}{p(y_0=1)}}{\frac{p(y_1 = 0|x_1,g_1)}{p(y_1=0)} \frac{p(y_0 = 1|x_0,g_0)}{p(y_0=1)}} \notag \\
    &= \frac{\group{\rho}(x_1,g_1) \paren*{1-\group{\rho}(x_0,g_0)}}{\paren*{1-\group{\rho}(x_1,g_1)} \group{\rho}(x_0,g_0)}\notag \\
   &= \frac{\oddsRatio\paren*{\alpha^{g_1}_{\pi(x_1)},\alpha_{\pi(x_1)}}r(x_1)}{ \oddsRatio\paren*{\alpha^{g_0}_{\pi(x_0)},\alpha_{\pi(x_0)}}r(x_0)} \notag \\
   &= \frac{\tau_r(x_1,x_0)}{\omega(g_0,\pi(x_0), g_1, \pi(x_1))}. \label{eq:tauomega}
\end{align}
Using the expression above, $\expt_{10}$ expectation of a finite valued function $\lambda(x_1,g_1,x_0,g_0)$, that vanishes to $0$ when $\tau_r(x_1,x_0)=\infty$ or $\omega(g_0,\pi(x_0), g_1, \pi(x_1))=0$,   can be expressed as an $\expt_{01}$ expectation as follows. \par

\begin{align*}
    \expt_{10}\brac*{\lambda(x_1,g_1,x_0,g_0)} &= \sum_{(g_1,g_0)\in \gset^2}\int_{(x_1,x_0)\in \xset^2} \lambda(x_1,g_1,x_0,g_0)  \group{f}_+(x_1,g_1)\group{f}_-(x_0,g_0)dx_1 dx_0 \\
    &= \sum_{(g_1,g_0)\in \gset^2}\int_{(x_1,x_0)\in \xset^2}\lambda(x_1,g_1,x_0,g_0)  \frac{\group{f}_+(x_1,g_1)\group{f}_-(x_0,g_0)}{\group{f}_+(x_0,g_0)\group{f}_-(x_1,g_1)}\group{f}_+(x_0,g_0)\group{f}_-(x_1,g_1)dx_1 dx_0 \tag{under certain conditions given below}\\
     &= \sum_{(g_1,g_0)\in \gset^2}\int_{(x_1,x_0)\in \xset^2} \lambda(x_1,g_1,x_0,g_0)  \frac{\tau_r(x_1,x_0)}{\omega(g_0,\pi(x_0), g_1, \pi(x_1))}\group{f}_+(x_0,g_0)\group{f}_-(x_1,g_1)dx_1 dx_0 \\
      &= \expt_{01}\brac*{\lambda(x_1,g_1,x_0,g_0)\frac{\tau_r(x_1,x_0)}{\omega(g_0,\pi(x_0), g_1, \pi(x_1))}}. \\
\end{align*}
The above derivation is valid if 
$$\group{f}_+(x_0,g_0)\group{f}_-(x_1,g_1) = 0 \Rightarrow  \lambda(x_1,g_1,x_0,g_0)\group{f}_+(x_1,g_1)\group{f}_-(x_0,g_0) = 0. $$
Next we show that if $\lambda(x_1,g_1,x_0,g_0)$ vanishes to $0$ when $\tau_r(x_1,x_0)=\infty$ or $\omega(g_0,\pi(x_0), g_1, \pi(x_1))=0$, then the above implication is indeed satisfied.
We present a proof by contradiction. Suppose $\group{f}_+(x_0,g_0)\group{f}_-(x_1,g_1) = 0$, but $\lambda(x_1,g_1,x_0,g_0)\group{f}_+(x_1,g_1)\group{f}_-(x_0,g_0) \neq 0.$ Thus from Eq. \ref{eq:tauomega}, $\frac{\tau_r(x_1,x_0)}{\omega(g_0,\pi(x_0), g_1, \pi(x_1))} = \infty$, which is only possible if $\tau_r(x_1,x_0)=\infty$ or $\omega(g_0,\pi(x_0), g_1, \pi(x_1))=0$. Thus $\lambda(x_1,g_1,x_0,g_0)=0$ and consequently $\lambda(x_1,g_1,x_0,g_0)\group{f}_+(x_1,g_1)\group{f}_-(x_0,g_0) = 0$, which leads to the contradiction.\par 
\noindent Note that $\expt_{10}$ expectation can also be converted to $\expt_{01}$ expectation by swapping the variables; i.e., $\expt_{10}\brac*{\lambda(x_1,g_1,x_0,g_0)}=\expt_{01}\brac*{\lambda(x_0,g_0,x_1,g_1)}$. Now, converting the $\expt_{10}$ expectations on the first three lines of the RHS of Eq. \ref{eq:expt} to $\expt_{01}$ expectations, using the two rules derived above, 

\begin{align*}
   \AUC\paren*{\group{\rho}}   %
    &= \AUC\paren*{\rho} + \expt_{01}\brac*{\Ind{\paren*{\omega^{01}, 1}}{\tau_r^{10}} \times\Ind{(0,1)}{\omega^{01}} \times \nicefrac{\tau_r^{10}}{\omega^{01}} } - \expt_{01}\brac*{\Ind{\paren*{1,\omega^{10}}}{\tau_r^{01}} \times \Ind{\paren*{1,\infty}}{\omega^{10}}}\notag\\
     & \quad +\frac{1}{2} \paren*{\expt_{01}\brac*{\dirac{ \omega^{01}}{\tau_r^{10}} \times\Ind{(0,1)}{\omega^{01}} \times \nicefrac{\tau_r^{10}}{\omega^{01}}} -\expt_{01}\brac*{\dirac{ \omega^{10}}{\tau_r^{01}} \times\Ind{(1,\infty)}{\omega^{10}}}} 
    \notag\\
    & \quad +\frac{1}{2} \paren*{\expt_{01}\brac*{\dirac{1}{\tau_r^{10} }\times\Ind{(0,1)}{\omega^{01}} \times \nicefrac{\tau_r^{10}}{\omega^{01}}}   - \expt_{01}\brac*{\dirac{1}{\tau_r^{01} }\times\Ind{(1,\infty)}{\omega^{10}}}} \notag\\
    &\quad + \expt_{10}\brac*{\Ind{\paren*{0, 1}}{\tau_r^{10}}\times \dirac{0}{\omega^{01}}}  +\frac{1}{2} \expt_{10}\paren*{\dirac{1}{\tau_r^{10}}\times \dirac{0}{\omega^{01}}}
\end{align*}
Next, observing that $\tau_r^{01} = \nicefrac{1}{\tau_r^{10}}$ and $\omega^{10} = \nicefrac{1}{\omega^{01}}$,
\begin{align*}
   \AUC\paren*{\group{\rho}}   %
    &= \AUC\paren*{\rho} + \expt_{01}\brac*{\Ind{\paren*{\omega^{01}, 1}}{\tau_r^{10}} \times\Ind{(0,1)}{\omega^{01}} \times \nicefrac{\tau_r^{10}}{\omega^{01}} } - \expt_{01}\brac*{\Ind{\paren*{\omega^{01},1}}{\tau_r^{10}} \times \Ind{(0,1)}{\omega^{01}}}\notag\\
    & \quad +\frac{1}{2} \paren*{\expt_{01}\brac*{\dirac{ \omega^{01}}{\tau_r^{10}} \times\Ind{(0,1)}{\omega^{01}} \times \nicefrac{\tau_r^{10}}{\omega^{01}}} -\expt_{01}\brac*{\dirac{ \omega^{01}}{\tau_r^{10}} \times\Ind{(0,1)}{\omega^{01}}}} 
    \notag\\
    & \quad +\frac{1}{2} \paren*{\expt_{01}\brac*{\dirac{1}{\tau_r^{10} }\times\Ind{(0,1)}{\omega^{01}} \times \nicefrac{\tau_r^{10}}{\omega^{01}}}   - \expt_{01}\brac*{\dirac{1}{\tau_r^{10} }\times\Ind{(0,1)}{\omega^{01}}}} \notag\\
 &\quad + \expt_{10}\brac*{\Ind{\paren*{0, 1}}{\tau_r^{10}}\times \dirac{0}{\omega^{01}}}  +\frac{1}{2} \expt_{10}\paren*{\dirac{1}{\tau_r^{10}}\times \dirac{0}{\omega^{01}}}\\
  &= \AUC\paren*{\rho} + \expt_{01}\brac*{\paren*{\Ind{\paren*{\omega^{01}, 1}}{\tau_r^{10}} + \frac{1}{2} \dirac{\omega^{01}}{\tau_r^{10}}+ \frac{1}{2} \dirac{1}{\tau_r^{10}}}\times\Ind{(0,1)}{\omega^{01}}\times \paren*{\nicefrac{\tau_r^{10}}{\omega^{01}} -1}}\\
  &\quad + \expt_{10}\brac*{\paren*{\Ind{\paren*{0, 1}}{\tau_r^{10}}+ \frac{1}{2}\dirac{1}{\tau_r^{10}}} \times \dirac{0}{\omega^{01}}}\\
   &= \AUC\paren*{\rho} + \expt_{01}\brac*{\paren*{\Ind{\paren*{\omega^{01}, 1}}{\tau_r^{10}} + \frac{1}{2} \dirac{1}{\tau_r^{10}}}\times\Ind{(0,1)}{\omega^{01}}\times \paren*{\nicefrac{\tau_r^{10}}{\omega^{01}} -1}}\\
 &\quad + \expt_{10}\brac*{\paren*{\Ind{\paren*{0, 1}}{\tau_r^{10}}+ \frac{1}{2}\dirac{1}{\tau_r^{10}}} \times \dirac{0}{\omega^{01}}} \tag{because $\dirac{\omega^{01}}{\tau_r^{10}}=1 \Rightarrow \nicefrac{\tau_r^{10}}{\omega^{01}} =1$ }
  \end{align*}
  Finally, again converting $\expt_{01}$ expectation in line 1 to $\expt_{10}$ by swapping the variables,
\begin{align*}
  \AUC\paren*{\group{\rho}} 
  &= \AUC\paren*{\rho} + \expt_{10}\brac*{\paren*{\Ind{\paren*{\omega^{10}, 1}}{\tau_r^{01}} + \frac{1}{2} \dirac{1}{\tau_r^{01}}}\times\Ind{(0,1)}{\omega^{10}}\times \paren*{\nicefrac{\tau_r^{01}}{\omega^{10}} -1}} \notag\\
    &\quad + \expt_{10}\brac*{\paren*{\Ind{\paren*{0, 1}}{\tau_r^{10}}+ \frac{1}{2}\dirac{1}{\tau_r^{10}}} \times \dirac{0}{\omega^{01}}}
\end{align*}
\end{proof}
\end{thm}

\begin{thm}
For $\lbl{\rho}(x) = \lbl{p}(y=1|x)$, $\lbl{\alpha}_{k} = \lbl{p}(y=1|\pi =k)$ and $\alpha^g_{k} = p(y=1|g,\pi=k)$,
$$\group{\rho}(x,g) = \paren*{1+ \oddsRatio\paren*{\lbl{\alpha}_{\pi(x)}, \alpha^g_{\pi(x)}}  \frac{1-\lbl{\rho}(x)}{\lbl{\rho}(x)}}^{-1}.$$
\begin{proof}
\begin{align*}
    \frac{\lbl{\rho}(x)}{1-\lbl{\rho}(x)} &= \frac{\tilde{p}(y=1|x)}{1-\tilde{p}(y=1|x)}\\
    &= \frac{\tilde{p}(y=1|x,\pi(x))}{1-\tilde{p}(y=1|x,\pi(x))}\\
    &= \frac{\tilde{p}(y=1|\pi(x))\tilde{p}(x|y=1,\pi(x))}{\tilde{p}(y=0|\pi(x))\tilde{p}(x|y=0,\pi(x))}\\
    &= \frac{\tilde{\alpha}_{\pi(x)}p(x|y=1,\pi(x))}{\paren*{1-\tilde{\alpha}_{\pi(x)}}p(x|y=0,\pi(x))} \tag{due to \cpccInv\ between labeled and unlabeled data}\\
     &= \oddsRatio\paren*{\tilde{\alpha}_{\pi(x)}, \alpha^g_{\pi(x)}}\frac{\alpha^g_{\pi(x)}p(x|y=1,\pi(x),g)}{\paren*{1-\alpha^g_{\pi(x)}}p(x|y=0,\pi(x),g)} \tag{due to \cpccInv\ across groups}\\
      &= \oddsRatio\paren*{\tilde{\alpha}_{\pi(x)}, \alpha^g_{\pi(x)}}\frac{p(y=1,x|\pi(x),g)}{p(y=0,x|\pi(x),g)}\\
      &= \oddsRatio\paren*{\tilde{\alpha}_{\pi(x)}, \alpha^g_{\pi(x)}}\frac{p(y=1|x,\pi(x),g)}{p(y=0|x,\pi(x),g)} \\
       &= \oddsRatio\paren*{\tilde{\alpha}_{\pi(x)}, \alpha^g_{\pi(x)}}\frac{p(y=1|x,g)}{p(y=0|x,g)} \\
       &= \oddsRatio\paren*{\tilde{\alpha}_{\pi(x)}, \alpha^g_{\pi(x)}}\frac{\group{\rho}(x,g)}{1-\group{\rho}(x,g)} \\
\end{align*}
Rearranging the terms,
\begin{align*}
     &\frac{\group{\rho}(x,g)}{1-\group{\rho}(x,g)} = \frac{1}{\oddsRatio\paren*{\tilde{\alpha}_{\pi(x)}, \alpha^g_{\pi(x)}} \frac{1-\lbl{\rho}(x)}{\lbl{\rho}(x)}} \\
     & \quad \Leftrightarrow \group{\rho}(x,g) = \frac{1}{1+\oddsRatio\paren*{\tilde{\alpha}_{\pi(x)}, \alpha^g_{\pi(x)}} \frac{1-\lbl{\rho}(x)}{\lbl{\rho}(x)}}. \\
\end{align*}
\end{proof}
\end{thm}

\FloatBarrier 
\section{Real-World Datasets }
Table \ref{tab:realDataStats} denotes the number of instances and dimensionality of the feature space for the real-world datasets used in the experiments along with a description of a positive class label. The data was obtained from the folktables Python package which uses data from the American Community Survey (ACS) Public Use Microdata Sample (PUMS) dataset. The 2018 5-Year horizon personal survey responses were used to generate all real-world datasets.
\begin{table}
\centering
\begin{tabular}{lrrr}
Dataset & $n$ & $d$ & Positive Class\\
\hline
Income & 8,116,931 & 10 & Person's income $>$ \$50,000 \\
Employment & 15,949,380 & 16 & Person is employed \\
IPR & 15,949,380 & 20 & Person's poverty status value $< 250$ \\
\end{tabular}
\caption{\label{tab:realDataStats}Number of instances ($n$) and dimensionality ($d$) for the real-world datasets used in the experiments along with a description of a positive target value}
\label{tab:dataset}
\end{table}

\section{Experiments on Amazon Reviews Dataset}

Our method was additionally evaluated on datasets generated from The Multilingual Amazon Reviews Dataset \cite{marc_reviews}. The classification task was to predict whether a given Amazon product review text corresponds to a product rating of more than 3 stars. The product category was used to partition reviews into groups. A 384-dimensional embedding was generated for each product review using the "all-MiniLM-L6-v2" language model \cite{wang2020minilm}. Datasets were generated following the same steps used for generating ACS datasets in settings 1 and 2. Table \ref{tab:amazon} denotes the AUC calculated on held-out tests, averaged over at least 10 iterations.

\begin{table}
\centering
\begin{tabular}{lrr}
Method &  Setting 1 &  Setting 2 \\
\hline
CORAL              &      0.744 &      0.709 \\
Global             &      0.745 &      0.718 \\
Group-Aware Global &      0.745 &      0.674 \\
Cluster Global     &      0.759 &      0.745 \\
Label Shift        &      0.695 &      0.702 \\
Our Method         &      0.706 &      0.751 \\
\hline
True Clustering               &      0.696 &      0.791 \\
\end{tabular}
\caption{Average AUC calculated on the held-out test set for Amazon datasets in settings 1 and 2.}
\label{tab:amazon}
\end{table}
\end{document}